\documentclass[llncsdoc,english]{llncs}
\usepackage[T1]{fontenc}
\usepackage[latin1]{inputenc}
\usepackage{graphicx}
\usepackage{amsmath}
\usepackage[numbers]{natbib}
\usepackage[]{algorithm2e}
\makeatletter

\renewcommand\bibsection%
{
  \section*{\refname
    \@mkboth{\MakeUppercase{\refname}}{\MakeUppercase{\refname}}}
}

\usepackage{babel}
\makeatother

\begin{document}

\vfill{}
\title{\textbf{Language Models with Pre-Trained (GloVe) Word Embeddings}} 
\vfill{}

\author{ Victor Makarenkov, Lior Rokach, Bracha Shapira 
\mbox{}\\
\email{makarenk@post.bgu.ac.il, liorrk@bgu.ac.il, bshapira@bgu.ac.il}
}

\institute{
Department of Software and Information Systems Engineering\\
Ben-Gurion University of the Negev
}
 
\maketitle

\section{Introduction}

In this work we present a step-by-step implementation of training a Language Model (LM) , using Recurrent Neural Network (RNN) and pre-trained GloVe word embeddings, introduced by Pennigton et al. in  \cite{glove2014}. The implementation is following the general idea of training RNNs for LM tasks presented in \cite{DBLP:journals/corr/ZarembaSV14} , but is rather using Gated Recurrent Unit (GRU) \cite{DBLP:journals/corr/ChoMGBSB14} for a memory cell, and not the more commonly used LSTM \cite{LSTM}. The implementation presented is based on using \texttt{keras}\footnote{https://keras.io/} \cite{keras}.

\section{Motivation}

Language Modeling is an important task in many Natural Language Processing (NLP) applications. These application include clustering, information retrieval, machine translation, spelling and grammatical errors correction. In general, a language model defined as a function that puts a probability measure over strings drawn from some vocabulary. In this work we consider a RNN based language model task, which aims at predicting the next \textit{n-th} word in a sequence, given the previous $n-1$ words. Put otherwise, finding the word with maximum value for $P(w_n|w_1,...,w_{n-1})$ . The $n$ parameter is the $ContextWindowSize$ argument in the algorithm described further.\\
To maximize the effectiveness and performance of the model we use word embeddings into a continuous vector space. The model of embedding we use is the GloVe \cite{glove2014} model, with dimensionality size equal to 300 or 50. We use both pre-trained\footnote{downloaded from: http://nlp.stanford.edu/projects/glove/}  on 42 billion tokens and 1.9 million vocabulary size, and specifically trained for this work vector models, which we trained on SIGIR-2015 and ICTIR-2015 conferences' proceedings.\\
The model itself is trained as a RNN, with internal GRU for memorizing the prior sequence of words. It was shown lately, that RNNs outperform most language modeling based tasks \cite{DBLP:journals/corr/ZarembaSV14, DBLP:journals/corr/ChoMGBSB14} when tuned and trained correctly.

\section{Short Description}
In this work we use 300-dimensional and 50-dimensional, GloVe word embeddings. In order to embed the words in a vector space, GloVe model is trained by examining word co-occurrence matrix $X_{ij}$ within a huge text corpus. Despite the huge size of the Common Crawl corpus, some words may not exist with the embeddings, so we set these words to random vectors, and use the same embeddings consistently if we encounter the same unseen word again in the text. The RNN is further trained to predict the next word in its embedding form, that is, predicts the next n-dimensional vector, given the $ContextWindowSize$ previous words. We divide the $TextFile$ into 70\% and 30\% for training and testing purposes.

\section{Pseudo Code}

\begin{algorithm}[H]

\SetKwData{Left}{left}\SetKwData{This}{this}\SetKwData{Up}{up}
\SetKwFunction{Union}{Union}\SetKwFunction{FindCompress}{FindCompress}
\SetKwInOut{Input}{input}\SetKwInOut{Output}{output}
 \Input{Input: glove-vectors : Pre-Trained-Word-Embeddings, Text-File, ContextWindowSize=10, hidden-unites=300}
 \Output{A Language Model trained on Text-File with word-embeddings representation}
 
 \For{ $w \in$  Text-File }{
	\If {$w \in$ OOV-file}{
		tokenized-file.append(OOV-file.get-vector(w))
	}
	\If {$w \in$ glove-vectors}{
		tokenized-file.append(glove-vectors.get-vector(w))
	}
	\Else {
		vector $\leftarrow$ random-vector() \\		
		tokenized-file.append(vector) \\
		OOV-file.append(vector)
	}
}	

NN $\leftarrow$ CreateSingleHiddenLayerDenseRNN(unit=GRU, inputs=300, outputs=300, hidden-unites)\\
NN.setDropout(0.8)\\
NN.setActivationFunction(Linear)\\
NN.setLossFunction(MSE)\\
NN.setOptimizer(rmsprop)\\
$X_{train} \leftarrow$ tokenized-file.getInterval(0.0,0.7)\\
$X_{test} \leftarrow$ tokenized-file.getInterval(0.7,1.0)\\
$Y_{train} \leftarrow X_{train}.Shift(ContextWindowSize)$\\
$Y_{test} \leftarrow X_{test}.Shift(ContextWindowSize)$\\

NN.Fit($X_{train}, Y_{train}$)\\
NN.Predict($X_{test}, Y_{test}$)\\

 \caption{Training a language model on word embeddings}
\end{algorithm}

\section{Detailed Explanation} \label{section:explain}

As stated earlier, GloVe model is trained by examining word co-occurrence matrix of two words $i$ and $j$: $X_{ij}$ within a huge text corpus. While training the main idea is stating that ${w_i}^Tw_j+b_i+b_j=log(X_{ij})$ where $w_i$ and $w_j$ are the trained vectors, $b_i$ and $b_j$ are the scalar bias terms associated with words $i$ and $j$. The most important parts of the training process in GloVe are: 1) A weighting function $f$ for elimination of very common words (like stop words) which add noise and not overweighted, 2) rare words are not overweighted 3) the co-occurrence strength, when modeled as a distance, should be smoothed with a $log$ function. Thus, the final loss function for a GloVe model is $J= \Sigma_{i,j \in V}f(X_{ij})({w_i}^Tw_j+b_i+b_j-log(X_{ij}))^2 $ where $V$ is a complete vocabulary, and $f(x)=(x/x_{max})^\alpha$ if $x < x_{max}$, and $f(x)=1$ otherwise. The model that is used in this work was trained with $x_{max}=100$ and $\alpha=0.75$.

\begin{figure}[h]
    \centering
    \includegraphics[width=1.0\textwidth]{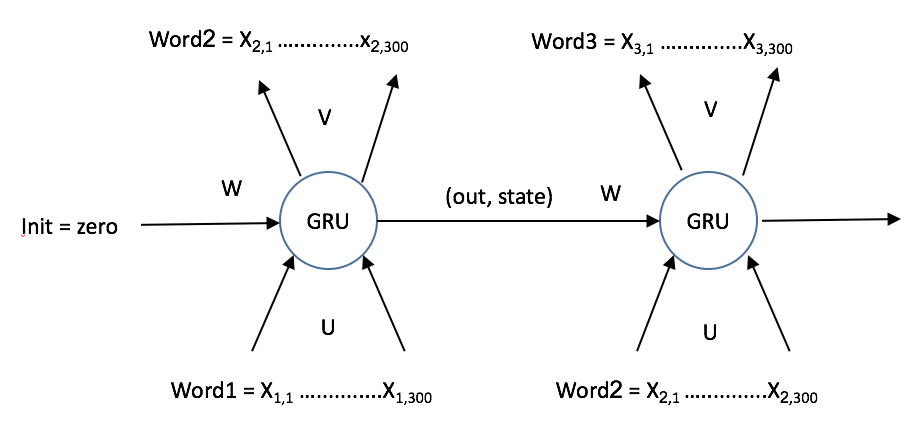}
    \caption{A general architecture of training an unfolded RNN with 1-sized window based shifted labeled data}
    \label{fig:RNN}
\end{figure}

Training of the RNN is somewhat blurred between supervised and unsupervised techniques. That is, no extra labeled data is given, but part of the input is used as labels. In this \textit{unfolded} training paradigm, which is illustrated on Figure \ref{fig:RNN}, the output is \textit{shifted} in a way to create a labels for the input train dataset. In this way the RNN can actually learn to predict a next word (vector) in a sequence.

\section{Evaluation}

\subsection{Pre-trained Vector Models}

\begin{figure}[h]
    \centering
    \includegraphics[width=0.5\textwidth]{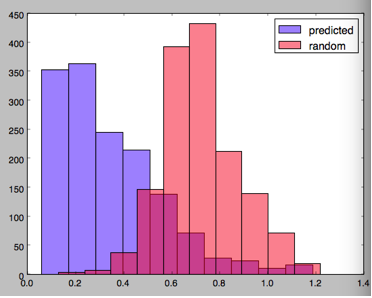}
    \caption{Two distributions of the predicted next word vector errors. The left is the result of predicted by RNN errors, and the righ is predicted by random. The RNN in the model was trained with 30 hidden GRU units. It took 300 iterations (epochs) on the data to achieve these results.}
    \label{fig:hidden30}
\end{figure}

In order to evaluate our implementation of the language model, we train several different language models and compare the predicted error distribution with a random word prediction. The error is measured with a cosine distance\footnote{implemented on python, with scipy package} between two vectors: $1- \frac {\pmb x \cdot \pmb y}{||\pmb x|| \cdot ||\pmb y||}$. On figure \ref{fig:hidden30} we see the error distribution of the RNN with 30 hidden units. The training was performed on 5000 tokens long text file, the first entry at English wikipedia, \textit{Anarchism}.

The machine that was used to run the evaluation had the following characteristics: 1.7 GHz, Core i7 with 8 GB memory, OS X version 10.11.13.

The time it took to train the model with 30 epochs was 125 seconds . The time took to make the predictions on a test set is 0.5 seconds.

\begin{figure}[h]
    \centering
    \includegraphics[width=0.5\textwidth]{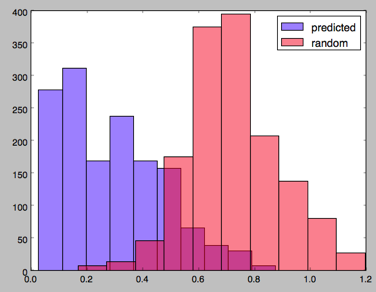}
    \caption{Two distributions of the predicted next word vector errors. The left is the result of predicted by RNN errors, and the righ is predicted by random. The RNN in the model was trained with 300 hidden GRU units. It took 300 iterations (epochs) on the data to achieve these results.}
    \label{fig:hidden300}
\end{figure}

On figure \ref{fig:hidden300} we see the error distribution of the RNN with 300 hidden units
The time it took to train the model with \textbf{300} epochs was 1298 seconds . The time took to make the predictions on a test set is 0.49 seconds.

\subsection{Self Trained Vector Model}

\begin{figure}[h]
    \centering
    \includegraphics[width=0.5\textwidth]{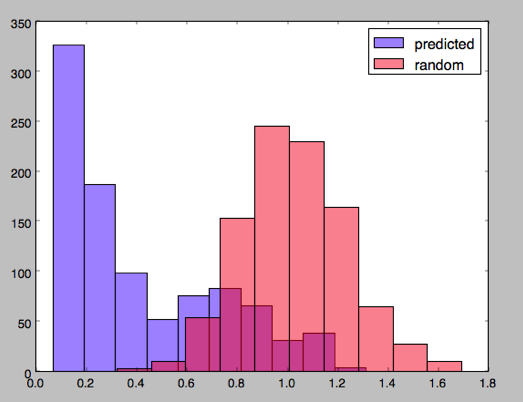}
    \caption{Two distributions of the predicted next word vector errors. The left is the result of predicted by RNN errors, and the righ is predicted by random. The RNN in the model was trained with 50 hidden GRU units. It took 50 iterations (epochs) on the data to achieve these results.}
    \label{fig:sigir}
\end{figure}

In addition, in order to further evaluate the current approach, we specifically trained a narrow domain-specific, vector model. We used ICTIR-2015 and SIGIR-2015 conferences proceedings as a corpora, and produced 50-dimensional vectors. The \textit{vector model} is based on 1,500,000 tokens total, and resulted in 17,000 long vocabulary. The \textit{language model} for the evaluation was built on a paper published in the ICTIR-2015 proceedings \cite{qpp2015}. Consider figure \ref{fig:sigir}. The predicted words' errors distribution differs even more than in the general case, where the vectors were trained on the general Common-Crawl corpora. That is, the performance of the language model, for the task of word prediction is higher.

The time took to train this model is 10 seconds. The time took to compute the predictions for evaluations is 0.04 seconds. 
\section{Instructions for running the code}
The implementation of the model training was written in this work in the Python language, version 2.7. The library that was used is Keras, which in the course of this implementation was based on \textit{Theano} framework. Instead of Theano, the Google's \textit{Tensorflow} can be also used behind the scenes of the Keras in this implementation.
In order to train the model yourself, you need to follow the next steps: 
\begin{enumerate}
	\item Download pre-trained GloVe vectors from \texttt{http://nlp.stanford.edu/projects/glove/}
	\item Obtain a text to train the model on. In our example we use a Wikipdia \textit{Anarchism} entry.
	\item Open and adjust the LM\_RNN\_GloVe.py file parameters inside the main function:
	\begin{enumerate}
	    	\item file\_2\_tokenize\_name (example = "/Users/macbook/corpora/text2tokenize.txt")
            	\item tokenized\_file\_name (example = "/Users/macbook/corpora/tokenized2vectors.txt")
            	\item glove\_vectors\_file\_name (example = "/Users/macbook/corpora/glove.42B.300d.txt")
       		\item extra\_vocab\_filename  (example = "/Users/macbook/corpora/extra\_vocab.txt"). This argument has also a default value in the \texttt{get\_vector} function
	\end{enumerate}
	\item Run the following methods:
	\begin{enumerate}
		\item tokenize\_file\_to\_vectors(glove\_vectors\_file\_name, file\_2\_tokenize\_name, tokenized\_file\_name)
    	        \item run\_experiment(tokenized\_file\_name)
	\end{enumerate}
\end{enumerate}

\section{Discussion}

In this work we implemented and tested the training of a LM based on RNN. To emphasize the strength of such an approach, we have chosen one of the most powerful and prominent techniques for word embeddings - the GloVe embeddings. Although there are other approaches, such as the popular \textit{word2vec} \cite{word2vec} technique, the GloVe embeddings was shown to outperform it on several tasks \cite{glove2014}, partially because of the reasons described in section \ref{section:explain}. By training the model with two different settings, one of which is order of magnitude more complex than the other we show the power of such LM. The distributions shown on figures \ref{fig:hidden30} and \ref{fig:hidden300} clearly indicate much smaller error on the task of next word prediction.

The main limitation of this implementation is the fixed window size, of the prefix in the LM. This approach does not fully show the full power of RNN-based LM. For dynamic size prefix LM please consider the DyNet \cite{dynet} package for example. DyNet supports a dynamic computation graph and shares the learned parameters across multiple variable length instances during the training.

\section{The source at GitHub}
The code was submitted publicly to the GitHub repository of the author, and is available under \textit{vicmak} username, \textit{proofseer} project\footnote{https://github.com/vicmak/ProofSeer}.
\newpage

\bibliographystyle{splncs}

\end{document}